\documentclass[runningheads]{llncs}
\usepackage{caption}
\usepackage[T1]{fontenc}
\usepackage{subcaption}
\usepackage{graphicx}
\usepackage{amsmath}
\usepackage{booktabs} 
\usepackage{xcolor}
\usepackage{authblk}
\usepackage{float}
\usepackage{multirow}
\begin{document}
\title{Passage Segmentation of
 Documents for Extractive Question Answering}
\author{Zuhong Liu\textsuperscript{1,2,3}\footnote{This work was done during an internship at BNP Paribas CIB, France} \and
Charles-Elie Simon\textsuperscript{2} \and
Fabien Caspani\textsuperscript{2} }
\authorrunning{Liu. et al.}
\institute{Ecole Polytechnique, France \and BNP Paribas CIB, France \and Paris Elite Institute of Technology, Shanghai Jiao Tong 
University, China \email{zuhong.liu@polytechnique.edu \{charles-elie.simon,fabien.caspani\}@bnpparibas.com}}
\maketitle              % typeset the header of the contribution
\begin{abstract}
Retrieval-Augmented Generation (RAG) has proven effective in open-domain question answering. However, the chunking process, which is essential to this pipeline, often receives insufficient attention relative to retrieval and synthesis components. This study emphasizes the critical role of chunking in improving the performance of both dense passage retrieval and the end-to-end RAG pipeline. We then introduce the Logits-Guided Multi-Granular Chunker (LGMGC), a novel framework that splits long documents into contextualized, self-contained chunks of varied granularity. Our experimental results, evaluated on two benchmark datasets, demonstrate that LGMGC not only improves the retrieval step but also outperforms existing chunking methods when integrated into a RAG pipeline.

\end{abstract}
\section{Introduction}
Open-Domain Question Answering (ODQA), which involves extracting a precise answer to a question from the content of a given document, has seen significant advancements with the advent of Retrieval-Augmented Generation (RAG) models \cite{lewis2021retrievalaugmentedgenerationknowledgeintensivenlp}. These models leverage large-scale pre-trained language models and retrieval systems to enhance the generation of accurate and contextually relevant answers. In a classical RAG pipeline, documents are initially split
 into independent chunks, and a retrieval process is
 applied to identify the relevant chunks to a given
 query. The retrieved chunks with the query are then passed as
 the prompt to the synthesizer LLM to get the desired response. Subsequent researches have concentrated on improving the two main aspects of RAG: retrieval \cite{karpukhin2020densepassageretrievalopendomain} \cite{izacard2022unsuperviseddenseinformationretrieval} \cite{wang2024improvingtextembeddingslarge} \cite{chen2023densexretrievalretrieval} and synthesis \cite{zhang2024raftadaptinglanguagemodel} \cite{wang2023learningfiltercontextretrievalaugmented} \cite{asai2023selfraglearningretrievegenerate}. However, few studies focus on investigating optimal solutions for document chunking and segmentation. The granularity and semantics intuitively play a significant role in precision during the retrieval stage. Besides, the absence of contextual information, as well as excessive irrelevant information within retrieved chunks can hinder the synthesizer LLM's ability to extract accurate key information despite the retriever's good performance. \cite{wang2023learningfiltercontextretrievalaugmented}

\begin{figure}[ht!]
    \vspace{2mm}
    \caption{Overview of the proposed method: Logits-Guided Multi-Granular Chunker. It utilizes the comprehension ability of LLMs to split documents into coherent parent chunks and split them into child chunks by Multi-Granular Module. The left figure shows how Multi-Granular Module works during the inference.}
    \centering
    \includegraphics[width=\textwidth]{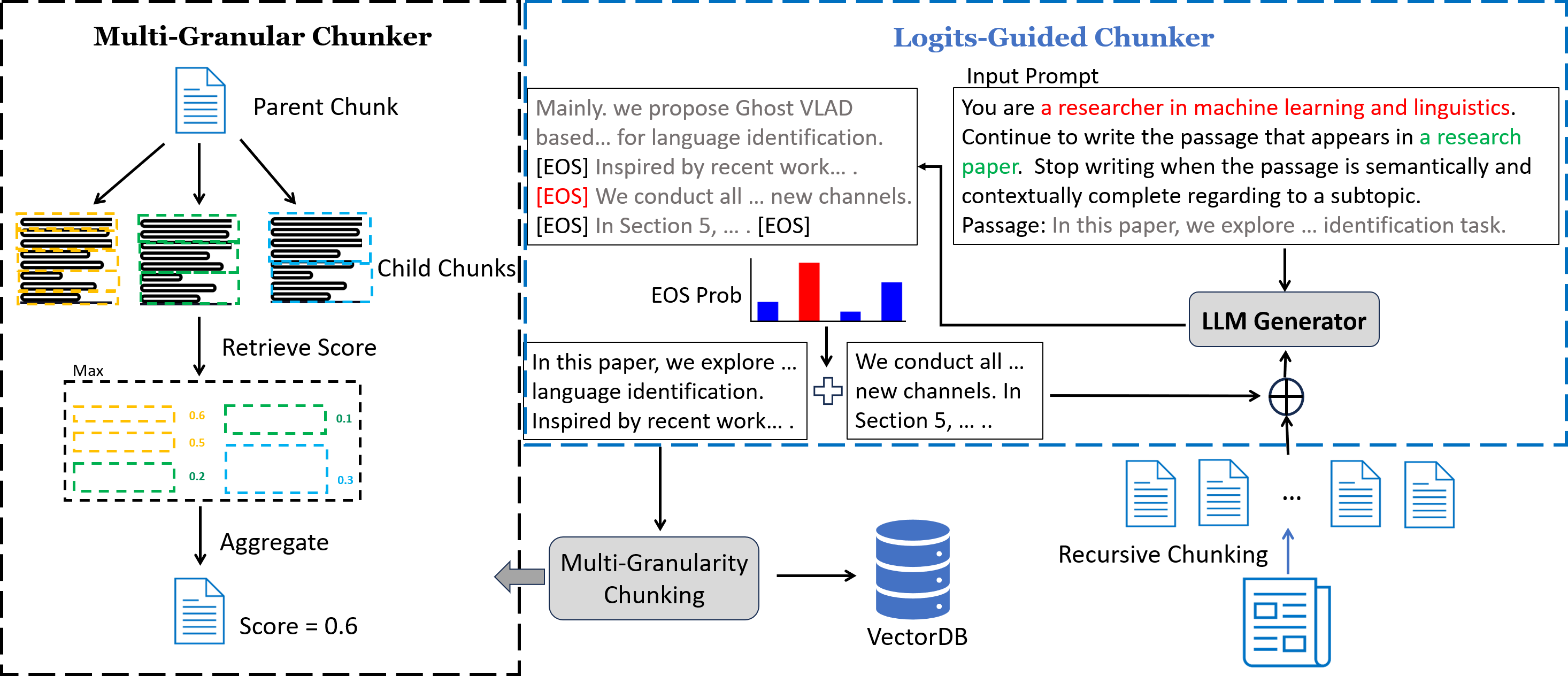}
    \label{fig:LGMGC}
    %\vspace{-6mm}
\end{figure}

In order to address the aforementioned challenges, we propose a new Logits-Guided Multi-Granular Chunker framework. It integrates two chunking modules: Logits-Guided Chunker and Multi-Granular Chunker within a unified framework shown in Figure \ref{fig:LGMGC}. The process begins by segmenting the documents into semantically and contextually coherent units, utilizing logits information derived from a smaller-scale LLM. Subsequently, these elementary units, referred to as parent chunks, are further divided into child chunks of varying granularity by the Multi-Granular Chunker in response to different types of query. Our results demonstrate that our approach performs favorably compared to current chunking methods on both passage retrieval and downstream question answering tasks.
\section{Related Work}
Several early works have explored chunk optimization for information retrieval. Recursive Chunking \cite{2024RecursiveChunking} segments text using a hierarchy of separators into units based on a predefined structure. Despite its simplicity, this approach may lack sufficient contextual information inside each chunk. To address this issue, Small2-Big \cite{2024Small2Big} introduces a method that retrieves information using small chunks and expands them into larger blocks with additional context for the LLM synthesizer. Semantic Chunking \cite{2024SemanticChunking} identifies breaking points based on significant embedding distances between sentences or passages, ensuring that the resulting chunks remain meaningful and coherent. Nevertheless, identifying optimal chunks for both retrieval and synthesis continues to be a challenging task.

Recent studies have also leveraged Large Language Models (LLMs) to extract text segments that are both contextually coherent and efficient for retrieval. \cite{chen2023densexretrievalretrieval} employs an LLM to transform documents into "propositions", atomic expressions that encapsulate distinct factoids, outperforming traditional passage or sentence-level chunking techniques. LumberChunker \cite{duarte2024lumberchunkerlongformnarrativedocument} autonomously identifies optimal segmentation points by iteratively feeding passages into an LLM. \cite{qian2024groundinglanguagemodelchunkingfree} presents a novel in-context retrieval approach for RAG systems, leveraging encoded hidden states of documents within LLMs to decode relevant passages without document chunking. These LLM-driven chunking strategies consistently demonstrate superior performance, highlighting the potential for further exploration of LLM-based passage segmentation. However, integrating large-scale LLMs into Retrieval-Augmented Generation (RAG) pipelines increases costs and processing times compared to traditional methods, particularly those using proprietary models like GPT-4 or Gemini-1.5, which also raise IT security concerns inside enterprises. In contrast to the aforementioned methods, our proposed LLM-based approach requires only logits information from a single forward pass, thereby making the chunking process more computationally efficient.

\section{Methodology} 
\subsection{Logits-Guided Chunker}
Given that Large Language Models (LLMs) exhibit a strong understanding of context, we propose an efficient method to utilize pretrained LLMs to determine the boundaries between complete semantic units within a text. Specifically, LLMs are capable enough of modelling the probability distribution of subsequent tokens conditioned on a given sequence of tokens. As a result, we can compute the probability $p_{[EOS]}$, the likelihood of the [EOS] (End of Sequence) token occurring after each sentence, conditioned on a prompt $\rho$ instructing the LLM to continue writing based on the provided sentence. Intuitively, a higher probability of the [EOS] token suggests that the LLM prefers to terminate generation at that point, indicating the semantic completeness of the preceding text. In practice, the input document is first transformed into a stream of fixed-size chunks of $\theta$ words segmented by recursive chunking. We then calculate the probability of the [EOS] token at the end of each sentence within the initial chunks and select the position with the highest probability as the break point. Formally, for a text chunk $c = s_1 \oplus s_2 \oplus \dots \oplus s_n$, the break point $b_{[EOS]}$ is defined as:
\begin{equation} b_{[EOS]} = {\arg\max}_{k \in \{1, \dots, n\}} p([EOS]|\rho, \underset{i=1}{\overset{k}{\oplus}}s_i)\end{equation}
where $\rho$ is the descriptive prompt and $\oplus$ denotes concatenation of sentences $s_i$.

The text preceding the break point is treated as an individual chunk, while the remaining content is concatenated with the next fixed-size chunk to form the input for subsequent iterations. This process is repeated by the chunker until the sequence length of the input text falls below a specified threshold. It is noteworthy that while the output of the Logits-Guided Chunker may vary in size, it is constrained by an upper bound.

\subsection{Multi-Granular Chunker}
\label{sec:mgc}
\cite{2024Small2Big} claims that decoupling text chunks used for retrieval and the text chunks used for synthesis could be advantageous. This approach uses smaller text chunks during the retrieval process and subsequently provides the larger text chunk to which the retrieved text belongs to the large language model. Building on this concept, we introduce a Multi-Granular Chunker Module that splits text into child chunks at various granularity levels. Initially, the document is divided into larger chunks of $\theta$ words by recursive chunking, termed parent chunks. Each parent chunk is then recursively subdivided into smaller child chunks of $\theta/2$ and $\theta/4$ words by recursive chunking, which ensures that a sentence is not divided between multiple chunks. During inference, the similarity score of a parent chunk is determined by the maximum score of its child chunks (including the parent chunk itself). The top k parent chunks with the highest scores are then passed to the LLM synthesizer for generating responses. 

\subsection{Logits-Guided Multi-Granular Chunker}
Both modules aim to enhance performance from the perspectives of semantics and granularity, offering complementary benefits. Consequently, we propose LGMGC, which utilizes the output of the Logits-Guided Chunker, characterized by $\theta$, as parent chunks and further subdivides these into child chunks with the same implementation as in Section \ref{sec:mgc}. 

\section{Experimental settings}
To evaluate the impact of our Logits-Guided Multi-Granular Chunker, we conduct experiments to see how it performs on retrieval and downstream QA tasks. 
% The objective of our experimental approach is to address the following research questions:

% - \textbf{RQ1: }Does LGMGC improve both retrieval and question answering performance ?

% - \textbf{RQ2: }How does each component of LGMGC contribute to the performance of retrieval and question answering ?
% Since each question's evidence consists of only one or two sentences, it is unlikely that we will split the evidence into separate chunks. Thus, there exists only one evidence chunk per question, regardless of the chunking strategy.

\noindent \textbf{Datasets and Metrics} To evaluate retrieval performance, we apply chunking strategies on GutenQA \cite{duarte2024lumberchunkerlongformnarrativedocument}, a benchmark with “needle in a haystack” type of
 question-answer pairs derived from narrative books. Since each question's evidence consists of only one or two sentences, the evidence will not be split between separate chunks. Following their evaluation scheme, we use both $\text{DCG}@k$ and $\text{Recall}@k$ as our evaluation metrics. We re-label the data due to the observation that several pieces of evidence are not directly present in the original text, likely resulting from the synthetic generation of the benchmark by LLM. This discrepancy reduces the matching rate and, consequently, diminishes the value of the metrics used to evaluate the correspondence between the retrieved chunks and the query. Specifically, for each evidence, we compute its ROUGE score with respect to each chunk and select the chunk with the highest score as the relevant chunk associated with the query.   
To evaluate end-to-end RAG performance, we use three single-document QA datasets from LongBench \cite{bai2024longbenchbilingualmultitaskbenchmark}: NarrativeQA \cite{kočiský2017narrativeqareadingcomprehensionchallenge}, QasperQA \cite{kočiský2017narrativeqareadingcomprehensionchallenge}, and MultifieldQA \cite{bai2024longbenchbilingualmultitaskbenchmark}. These tasks are designed for information extraction without requiring advanced reasoning. Consistent with \cite{bai2024longbenchbilingualmultitaskbenchmark}, we use the F1-score as the evaluation metric, formally defined as follows:
\begin{equation*}
\text{Precision} = \frac{|\text{BOW}(\text{pred}) \cap \text{BOW}(\text{gt})|}{|\text{BOW}(\text{pred})|}, \quad
\text{Recall} = \frac{|\text{BOW}(\text{pred}) \cap \text{BOW}(\text{gt})|}{|\text{BOW}(\text{gt})|}, \\
\end{equation*}
\begin{equation*}
F1 = \frac{2 \cdot \text{Precision} \cdot \text{Recall}}{\text{Precision} + \text{Recall}}.
\end{equation*}

where BOW(A) represents the bag of words of the predicted ($pred$) or the ground-truth answer ($gt$). $\cap$ represents the operation that takes the minimum frequency for each word appearing in both sets. 

\noindent
\textbf{Baselines} We evaluate our approach against several established methods: \textbf{Recursive Chunker}\cite{2024RecursiveChunking} and \textbf{Semantic Chunker} \cite{2024SemanticChunking} which are among the most widely adopted in the field. Additionally, for the passage retrieval task, we include the \textbf{Paragraph-Level Chunker} (denoted as Para Chunker) and \textbf{LumberChunker}, as introduced by \cite{duarte2024lumberchunkerlongformnarrativedocument}, as further baseline comparisons. Furthermore, we incorporate two submodules, the \textbf{Multi-Granular Chunker} (denoted as MG Chunker) and the \textbf{Logits-Guided Chunker} (denoted as LG Chunker), as baselines for our ablation studies. We employ a 8-bit quantized Llama3-8b for Logits-Guided Chunker and LGMGC. All strategies are evaluated across different chunk sizes $\theta$ (200, 300, 500 words) to assess their sensitivity to this hyper-parameter. 

\noindent \textbf{Implementation Details}
We utilize standalone retrievers and synthesizers with varying chunking strategies to assess their performance. For retrievers, we employ two dense retrievers: BGE-Large \cite{xiao2024cpackpackagedresourcesadvance} in passage retrieval and additionally E5-Large \cite{wang2024textembeddingsweaklysupervisedcontrastive} in downstream QA.
For synthesizers, we utilize both small-scale and large-scale LLMs, Llama3-8b-Instruct (4-bit quantized), and Llama3-70b-Instruct (fp16) \cite{dubey2024llama}. Final answers are generated via greedy search to minimize randomness. To have a fair comparison between methods depending on the chunk size, we set an upper bound of 1500 words for the input context length of synthesizers. The top k chunks are sequentially concatenated into context until it reaches the upper bound. For the baseline without retrieval, the document is truncated to 3500 words due to computational constraints.
%  \begin{align}
%      \text{Recall@}k & = \frac{1}{|Q|}\sum_{q \in Q}\frac{\#retr_{q,k}}{\#rel_q} \\
%      \text{DCG@}k & = \frac{1}{|Q|}\sum_{q \in Q}\sum_{i=1}^{k} \frac{2^{\text{rel}_i} - 1}{\log_2(i + 1)}
%  \end{align}
 
% where 

% \begin{itemize}
%     \item $rel_{i}$ is a binary variable indicating the relevance of the chunk to the question as determined by the annotation.
    
%     \item $\#rel_q$: the number of evidence texts for a certain question $q$.
%     \item $\#retr_{q,k}$: the number of relevant chunks corresponding to question $q$ in Top-k retrieved parent chunk.
% \end{itemize}
\section{Results}
\begin{table}[ht]
\caption{Results on GutenQA using BGE Large. Mean and standard deviation are calculated across chunk sizes. Chunking methods that do not take
 chunk size as hyper-parameter have no standard deviation. Best
 results marked in \textbf{bold}.}
\centering
\small % Reduce font size
\setlength{\tabcolsep}{5pt} % Adjust column spacing

\begin{subtable}[t]{\textwidth}
\vspace{2mm}
\centering
\setlength{\tabcolsep}{0.3mm}{
\begin{tabular}{lccccc}
\toprule
\multirow{2}{*}{Method}& \multicolumn{5}{c}{DCG@k} \\
\cmidrule(lr){2-6}
& 1 & 2 & 5 & 10 & 20 \\
\midrule
Recursive Chunker & $47.22 \pm \textcolor{gray}{2.47}$ & $56.68 \pm \textcolor{gray}{2.59}$ & $62.32 \pm \textcolor{gray}{2.68}$ & $64.54 \pm \textcolor{gray}{2.40}$ & $66.28 \pm \textcolor{gray}{2.19}$ \\
MG Chunker & $60.33 \pm \textcolor{gray}{3.81}$ & $69.80 \pm \textcolor{gray}{2.64}$ & $74.66 \pm \textcolor{gray}{2.62}$ & $76.58 \pm \textcolor{gray}{2.47}$ & $77.47 \pm \textcolor{gray}{2.42}$ \\
Semantic Chunker & $41.33 \pm \textcolor{gray}{2.42}$ & $48.34 \pm \textcolor{gray}{3.28}$ & $54.23 \pm \textcolor{gray}{3.73}$ & $56.58 \pm \textcolor{gray}{3.42}$ & $58.68 \pm \textcolor{gray}{2.97}$ \\
LG Chunker & $52.55 \pm \textcolor{gray}{2.28}$ & $60.40 \pm \textcolor{gray}{1.88}$ & $67.29 \pm \textcolor{gray}{1.63}$ & $69.14 \pm \textcolor{gray}{1.50}$ & $70.38 \pm \textcolor{gray}{1.41}$ \\
Para Chunker& $50.00 \pm \textcolor{gray}{0.00}$ & $57.57 \pm \textcolor{gray}{0.00}$ & $60.62 \pm \textcolor{gray}{0.00}$ & $61.73 \pm \textcolor{gray}{0.00}$ & $63.12 \pm \textcolor{gray}{0.00}$ \\
LumberChunker & $55.67 \pm \textcolor{gray}{0.00}$ & $63.87 \pm \textcolor{gray}{0.00}$ & $69.91 \pm \textcolor{gray}{0.00}$ & $72.12 \pm \textcolor{gray}{0.00}$ & $74.59 \pm \textcolor{gray}{0.00}$ \\
LGMGC & $\textbf{63.00} \pm \textcolor{gray}{1.36}$ & $\textbf{72.46} \pm \textcolor{gray}{1.28}$ & $\textbf{76.43} \pm \textcolor{gray}{0.73}$ & $\textbf{78.37} \pm \textcolor{gray}{0.65}$ & $\textbf{78.96} \pm \textcolor{gray}{0.78}$ \\
\bottomrule
\end{tabular}
}
\end{subtable}
\hspace{20pt} % Space between the two subtables
\begin{subtable}[t]{\textwidth}
\centering
\setlength{\tabcolsep}{0.3mm}{
\begin{tabular}{lccccc}
\toprule
& \multicolumn{5}{c}{Recall@k} \\
\cmidrule(lr){2-6}
& 1 & 2 & 5 & 10 & 20 \\
\midrule
Recursive Chunker & $47.22 \pm \textcolor{gray}{2.47}$ & $62.22 \pm \textcolor{gray}{2.68}$ & $74.67 \pm \textcolor{gray}{2.68}$ & $81.55 \pm \textcolor{gray}{1.91}$ & $88.45 \pm \textcolor{gray}{1.29}$ \\
MG Chunker & $60.33 \pm \textcolor{gray}{3.81}$ & $75.33 \pm \textcolor{gray}{1.96}$ & $86.00 \pm \textcolor{gray}{1.90}$ & $91.89 \pm \textcolor{gray}{1.55}$ & $\textbf{95.33} \pm \textcolor{gray}{1.96}$ \\
Semantic Chunker & $41.33 \pm \textcolor{gray}{2.42}$ & $52.78 \pm \textcolor{gray}{3.86}$ & $65.33 \pm \textcolor{gray}{4.77}$ & $72.55 \pm \textcolor{gray}{3.81}$ & $81.00 \pm \textcolor{gray}{2.05}$ \\
LG Chunker & $52.55 \pm \textcolor{gray}{2.28}$ & $65.00 \pm \textcolor{gray}{1.65}$ & $80.11 \pm \textcolor{gray}{1.13}$ & $85.89 \pm \textcolor{gray}{0.79}$ & $90.67 \pm \textcolor{gray}{0.54}$ \\
Para Chunker& $50.00 \pm \textcolor{gray}{0.00}$ & $62.00 \pm \textcolor{gray}{0.00}$ & $68.67 \pm \textcolor{gray}{0.00}$ & $72.00 \pm \textcolor{gray}{0.00}$ & $77.33 \pm \textcolor{gray}{0.00}$ \\
LumberChunker & $55.67 \pm \textcolor{gray}{0.00}$ & $68.67 \pm \textcolor{gray}{0.00}$ & $81.67 \pm \textcolor{gray}{0.00}$ & $88.33 \pm \textcolor{gray}{0.00}$ & $94.33 \pm \textcolor{gray}{0.00}$ \\
LGMGC & $\textbf{63.00} \pm \textcolor{gray}{1.36}$ & $\textbf{78.00} \pm \textcolor{gray}{1.25}$ & $\textbf{86.78} \pm \textcolor{gray}{0.32}$ & $\textbf{92.56} \pm \textcolor{gray}{0.16}$ & $94.89 \pm \textcolor{gray}{0.31}$ \\
\bottomrule
\end{tabular}
}
\end{subtable}
\label{tab:gutenqa}
%vspace{-7mm}
\end{table}
\subsection{Passage Retrieval}
As shown in Table \ref{tab:gutenqa}, the Logits-Guided Chunker consistently outperforms the Recursive Chunker, Semantic Chunker, and Paragraph-Level Chunker across various chunk sizes, indicating its superior ability to capture contextual coherence and produce independent, concentrated semantic chunks. While slightly less effective than the LumberChunker, the Logits-Guided Chunker is more cost-effective and simpler to deploy, avoiding recursive LLM API calls and supporting local implementation. The Multi-Granular Chunker also shows significant performance improvement. At the end, LGMGC, considering both coherence of text chunks and granularity for different questions, achieves optimal results. Additionally, we observe that LGMGC exhibits the smallest standard deviation across varying chunk sizes, indicating that it maintains consistent performance regardless of the chunk size $\theta$, thereby reducing the need for hyper-parameter tuning in real-world use cases. This characteristic enhances the method's applicability and efficiency, as it is more robust on data-specific requirements.
% The reduced performance at $\theta = 500$ is attributed to the inclusion of multiple subtopics in larger segments. Consequently, the Logits-Guided Chunker may partition the segment at the boundary of the second subtopic, leading to semantic dilution. Thus, smaller chunk sizes are generally more effective for both the Logits-Guided Chunker and LGMGC.
% \begin{figure}[htbp]
%     \centering
%     \caption{Results on GutenQA by BGE Large. Chunking methods that do not take chunk size as hyper-parameter are represented by a horizontal line.}
%     \label{fig:2x2layout}
%     \includegraphics[width=0.8\textwidth]{images/passage_retrieval.png}
% \end{figure}

\subsection{Open Domain QA}
Table \ref{tab:longbench} provides an overview of our results on LongBench. We show the score of the optimal chunk size (the one with the highest score among all chunk sizes) as the score for a given chunking method. The results first indicate that applying the RAG pipeline yields better performance compared to providing the entire document to the synthesizers. Regarding chunker performance, the results align with those from the passage retrieval evaluation. LGMGC demonstrates the highest performance on all three datasets when using the optimal chunk size across various retrievers and synthesizers. This suggests that it yields superior outcomes on downstream question answering compared to current baselines.
\begin{table*}[ht]
    \vspace{-8mm}
    \footnotesize
    \caption{Results on LongBench Single DocQA Tasks. We use F1-score as the evaluation metric.
    NQA, MQA, QAP represent respectively NarrativeQA, Multi-fieldQA, and QasperQA. The best results are in \textbf{bold}.}
    %\vspace{2mm}
    \centering
    \setlength{\tabcolsep}{1mm}{
    \begin{tabular}{cccccccccc}
        \toprule
        \multirow{2}{*}{Retriever} & \multirow{2}{*}{Chunking Method} & \multicolumn{4}{c}{Llama3-8b-4bit} & \multicolumn{4}{c}{Llama3-70B-fp16} \\
        \cmidrule(lr){3-6} \cmidrule(lr){7-10}
         & & NQA & MQA & QAP & Avg. & NQA & MQA & QAP & Avg. \\
        \midrule
        \multirow{3}{*}{BGE Large} & Recursive Chunker & 18.1 & 49.3 & 39.1 & 35.5 & 22.9 & 51.0 & 45.7 & 39.9 \\
         & Semantic Chunker & 19.0 & 42.7 & 39.6 & 33.8 & 22.9 & 44.7 & 41.8 & 36.5 \\
        & LGMGC & \textbf{19.6} & \textbf{50.1} & \textbf{43.5} & \textbf{37.7} & \textbf{24.2} & \textbf{53.0} & \textbf{50.7} & \textbf{42.6} \\
        \midrule
        \multirow{3}{*}{E5 Large} & Recursive Chunker & 19.6 & 48.6 & 40.1 & 36.1 & \textbf{26.7} & 52.7 & 48.7 & 42.7 \\
         & Semantic Chunker & 18.7 & 47.6 & 39.4 & 35.2 & 24.6 & 51.0 & 47.5 & 41.0 \\
        & LGMGC & \textbf{19.7} & \textbf{51.4} & \textbf{42.6} & \textbf{37.9} & 26.1 & \textbf{55.1} & \textbf{50.0} & \textbf{43.7} \\
        \midrule
        \multicolumn{2}{c}{w/o retrieval} & 19.6 & 43.7 & 40.7 & 34.7 & 22.3 & 48.2 & 42.2 & 37.6 \\
        \bottomrule
    \end{tabular}
    }
    \vspace{-10mm}
    \label{tab:longbench}
\end{table*}

\section{Conclusions}
    
In this study, we propose a novel chunking framework named the Logits-Guided Multi-Granular Chunker, which consists of two primary components. The Logits-Guided Chunker aims to partition the document into contextually coherent and self-contained parent chunks, thereby improving the effectiveness of information retrieval. Furthermore, the Multi-Granular Module can further subdivide these parent chunks into smaller chunks of varying granularity. Our experiments show that LGMGC performs favorably compared to mainstream chunking methods in both retrieval and downstream QA tasks. These findings highlight its potential for further research and practical applications. In the future, integrating more reasonable automatic evaluation metrics on QA tasks that are more aligned with human evaluation, such as \cite{ravi2024lynxopensourcehallucination} \cite{es2023ragasautomatedevaluationretrieval} may help people understand how the chunking strategy matters in a more comprehensive way.
% \begin{credits}
% \subsubsection{\ackname} A bold run-in heading in small font size at the end of the paper is
% used for general acknowledgments, for example: This study was funded
% \end{credits}
\bibliographystyle{splncs04}
\bibliography{samplepaper}
\end{document}